\pdfoutput=1
\documentclass{article}

    \usepackage[final]{neurips_2020}

\usepackage[utf8]{inputenc} % allow utf-8 input
\usepackage[T1]{fontenc}    % use 8-bit T1 fonts
\usepackage[hidelinks]{hyperref}      % hyperlinks
\usepackage{url}            % simple URL typesetting
\usepackage{booktabs}       % professional-quality tables
\usepackage{amsfonts}       % blackboard math symbols
\usepackage{nicefrac}       % compact symbols for 1/2, etc.
\usepackage{microtype}      % microtypography

\usepackage{amsmath,amsfonts,bm}

\def\eqref#1{equation~\ref{#1}}
\def\1{\bm{1}}

\DeclareMathAlphabet{\mathsfit}{\encodingdefault}{\sfdefault}{m}{sl}
\SetMathAlphabet{\mathsfit}{bold}{\encodingdefault}{\sfdefault}{bx}{n}

\DeclareMathOperator*{\argmin}{arg\,min}

\DeclareMathOperator{\Tr}{Tr}

\usepackage{graphicx}
\usepackage{amssymb}
\usepackage{amsmath}
\usepackage{threeparttable}
\usepackage[ruled,vlined, noend]{algorithm2e}
\usepackage{setspace}
\usepackage{algpseudocode}
\usepackage{cite}
\usepackage{url}
\usepackage{color}
\usepackage{bm}
\usepackage{footnote}
\usepackage{wrapfig}
\usepackage{multirow}
\usepackage[normalem]{ulem}
\usepackage{soul} % strickthrough
\usepackage{makecell}
\usepackage{subcaption}
\usepackage{authblk}

\newcommand{\xv        }{\boldsymbol x        }
\newcommand{\yv        }{\boldsymbol y        }

\newcommand{\uv        }{\boldsymbol u        }

\newcommand{\Mv        }{\boldsymbol M        }

\newcommand{\thetav    }{\boldsymbol \theta   }

\newcommand{\tBigX     }{\Tilde{            X}}

\newcommand{\txv       }{\Tilde{\boldsymbol x}}

\newcommand{\beginsupplement}{%
        \setcounter{table}{0}
        \renewcommand{\thetable}{S\arabic{table}}%
        \setcounter{figure}{0}
        \renewcommand{\thefigure}{S\arabic{figure}}%
        \setcounter{equation}{0}
        \renewcommand{\theequation}{S\arabic{equation}}%
        \setcounter{section}{0}
        \renewcommand{\thesection}{S\arabic{section}}  
        \setcounter{page}{1}
        \renewcommand{\thepage}{S\arabic{page}} 
        
     }

\title{Task-Agnostic Online Reinforcement Learning with an Infinite Mixture of Gaussian Processes}

\author{%
  \textbf{Mengdi Xu, Wenhao Ding, Jiacheng Zhu, Zuxin Liu, Baiming Chen, Ding Zhao
  }\\
  Carnegie Mellon University\\
  \texttt{\{mengdixu, wenhaod, jzhu4, zuxinl, baimingc, dingzhao\}@andrew.cmu.edu}
}
\begin{document}

\maketitle

% =====================
% modifications made in the camera-ready version.
% 1. add extra experiments as suggested by reviewers
% 2. add more algorithm description and design motivation
% 3. add derivation of equation 2 in supplementary material
% 4. correct typos in equations
% 5. clarify the concerns in the related work
% 6. describe the hyper-parameter selection in supplementary material
%================================ abstract ================================

\begin{abstract}

Continuously learning to solve unseen tasks with limited experience has been extensively pursued in meta-learning and continual learning, but with restricted assumptions such as accessible task distributions, independently and identically distributed tasks, and clear task delineations. However, real-world physical tasks frequently violate these assumptions, resulting in performance degradation. This paper proposes a continual online model-based reinforcement learning approach that does not require pre-training to solve task-agnostic problems with unknown task boundaries. We maintain a mixture of experts to handle nonstationarity, and represent each different type of dynamics with a Gaussian Process to efficiently leverage collected data and expressively model uncertainty. We propose a transition prior to account for the temporal dependencies in streaming data and update the mixture online via sequential variational inference. Our approach reliably handles the task distribution shift by generating new models for never-before-seen dynamics and reusing old models for previously seen dynamics. In experiments, our approach outperforms alternative methods in non-stationary tasks, including classic control with changing dynamics and decision making in different driving scenarios. Codes available at: \url{https://github.com/mxu34/mbrl-gpmm}.

%================================Version 4================================

% Continuously learning to solve unseen tasks with limited experience has been extensively pursued in meta-learning and continual learning, but with restricted assumptions such as accessible task distributions, independently and identically distributed tasks, and clear task delineations. 
% However, real-world physical tasks frequently violate these assumptions, resulting in performance degradation. 
% This paper proposes a continual online model-based reinforcement learning approach that does not require pre-training to solve realistic, task-agnostic problems with unknown task boundaries. 
% We maintain a mixture of experts to handle nonstationarity, and represent each different type of dynamics with a Gaussian Process to efficiently leverage prior knowledge and expressively model uncertainty. 
% We propose a transition prior to account for the temporal dependencies in streaming data and update the mixture online via sequential variational inference.
% Our approach reliably handles the task distribution shift by generating new models for never-before-seen dynamics and reusing old models for previously seen dynamics. 
% In experiments, our approach outperforms alternative methods in non-stationary tasks including classic control with changing dynamics and decision making in different driving scenarios.

\end{abstract}

%================================ Introduction ================================

\section{Introduction}
\label{sec:intro}
% key concepts: model agnostic/non-stationary, pretrain-free, temporal dependency, data distillation, online

Humans can quickly learn new tasks from just a handful of examples by preserving rich representations of experience~\citep{lake2015human}. Intelligent agents deployed in the real world require the same continual and quick learning ability to safely handle unknown tasks, such as navigation in new terrains and planning in dynamic traffic scenarios.
Such desiderata have been previously explored in meta-learning and continual learning. Meta-learning~\citep{smundsson2018meta, clavera2018learning} achieves quick adaptation and good generalization with learned inductive bias. It assumes that the tasks for training and testing are independently sampled from the same accessible distribution. Continual learning~\citep{chen2018lifelong, v.2018variational} aims to solve a sequence of tasks with clear task delineations while avoiding catastrophic forgetting.
% with the iid assumption released. 
Both communities favor Deep neural networks (DNNs) due to their strong function approximation capability but at the expense of data efficiency. 
These two communities are complementary, and their integration is explored in~\citep{jerfel2019reconciling}. 

However, real-world physical tasks frequently violate essential assumptions of the methods as mentioned above. One example is the autonomous agent navigation problem requiring interactions with surrounding agents. 
The autonomous agent sequentially encounters other agents that have substantially different behaviors (e.g., aggressive and conservative ones).
In this case, the mutual knowledge transfer in meta-learning algorithms may degrade the generalization performance~\citep{deleu2018effects}. 
The task distribution modeling these interactions is prohibitively complex to determine, which casts difficulties on the meta-training process with DNNs~\citep{nagabandi2018deep, vuorio2019multimodal}. 
Additionally, the boundaries of tasks required in most continual learning algorithms cannot feasibly be determined beforehand in an online learning setting. Although task-agnostic/task-free continual learning is explored in~\citep{aljundi2019task, Lee2020A}, the temporal dependencies of dynamics presented in a non-stationary robotics task are missed. For instance, two different dynamics models close together in time are likely to be related.

\begin{figure}
    \centering
    \includegraphics[width=0.95\textwidth]{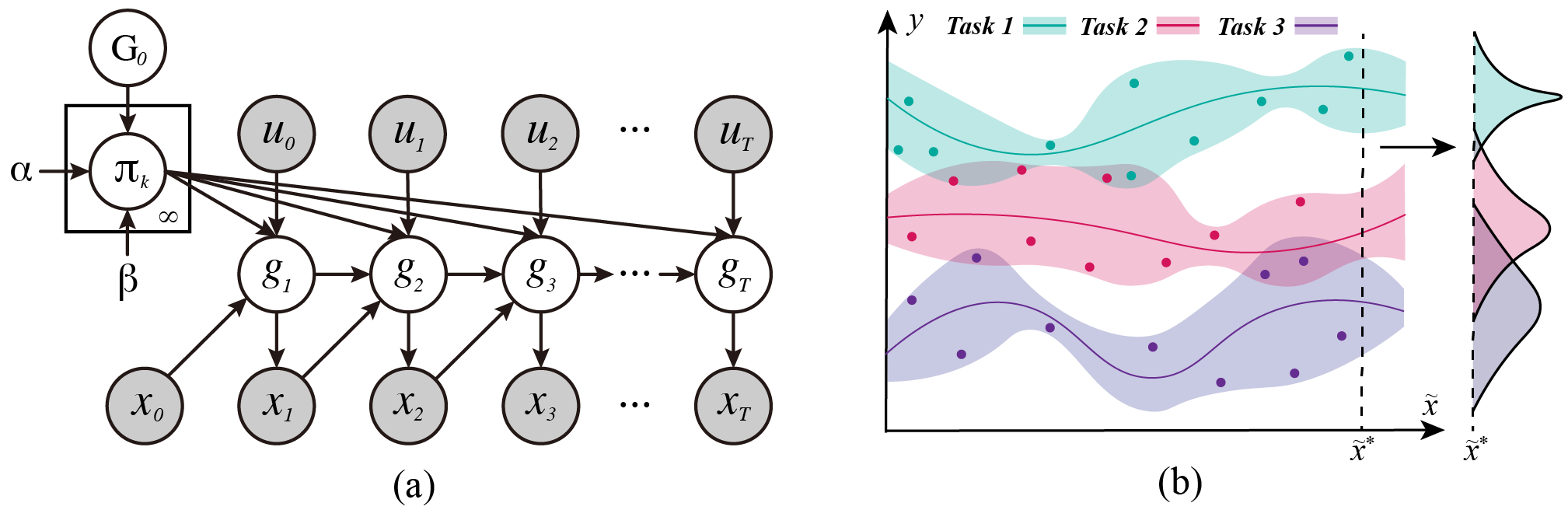}
    \caption{Method illustration. 
    \textbf{(a)} is a graphical representation of the proposed model-based RL with an infinite mixture as the dynamics model. $u_t$, $x_t$, and $g_t$ represent the action, state, and the dynamics model at time $t$, respectively. Parameters include the concentration parameter $\alpha$, the base distribution $G_0$, and the sticky parameter $\beta$. \textbf{(b)} visualizes the predictive distribution at a data point $\tilde{x}^*$.
    } 
    \label{fig:graphic}
\end{figure}

In this work, we aim to solve nonstationary online problems where the task boundaries and the number of tasks are unknown by proposing a model-based reinforcement learning (RL) method that does not require a pre-trained model.
Model-based methods~\citep{chua2018deep} are more data-efficient than model-free ones, and their performance heavily depends on the accuracy of the learned dynamics models.
Similar to expansion-based continual learning methods~\citep{jerfel2019reconciling, yoon2018lifelong}, we use an infinite mixture to model system dynamics, a graphical illustration of which is given in Figure~\ref{fig:graphic}~(a). 
It has the capacity to model an infinite number of dynamics, while the actual number is derived from the data. 
We represent each different type of dynamics with a Gaussian Process (GP)~\citep{rasmussen2003gaussian}
% or its neural network approximation, a Neural Process~\citep{pmlr-v80-garnelo18a, kim2018attentive}, 
to efficiently leverage collected data and expressively model uncertainty. 
A GP is more data-efficient than a DNN (as its predictive distribution is explicitly conditioned on the collected data) and thus enables fast adaptation to new tasks even without the use of a previously trained model.
With a mixture of GPs, the predictive distribution at a data point is multimodal, as shown in Figure~\ref{fig:graphic}~(b), with each mode representing a type of dynamics. By making predictions conditioned on the dynamics assignments, our method robustly handles dynamics that are dramatically different.

At each time step, our method either creates a new model for previously unseen dynamics or recalls an old model for encountered dynamics. After task recognition, the corresponding dynamics model parameters are updated via conjugate gradient~\citep{gardner2018gpytorch}.
% learned via Bayesian inference. 
Considering that RL agents collect experience in a streaming manner, we learn the mixture with sequential variational inference~\citep{lin2013online} that is suitable for the online setting.
% , which starts with the base measure and gradually transforms it into an approximate posterior. 
To account for the temporal dependencies of dynamics, we propose a transition prior that stems from the Dirichlet Process (DP) prior to improve task shift detection.
% , as detailed in Section~\ref{Exp:SwitchingDetection}. 
We select representative data points for each type of dynamics by optimizing a variational objective widely used in the Sparse GP literature~\citep{titsias2009variational}.
% To keep a reasonably sized memory buffer for each task, we select representative data points in order to approximate posterior distributions by optimizing a variational objective widely used in the Sparse GP literature~\citep{titsias2009variational}.
We demonstrate the capability of task recognition and quick task adaptation of our approach in non-stationary $\mathtt{Cartpole}$-$\mathtt{Swing Up}$, $\mathtt{HalfCheetah}$ and $\mathtt{Highway}$-$\mathtt{Intersection}$ environments.

% \begin{wrapfigure}{l}{0.5\textwidth}
% \end{wrapfigure}

%================================ Related Work ================================

\section{Related Work}
\label{sec:related work}

Meta-learning algorithms in general consist of a base (quick) learner and a meta (slow) learner~\citep{duan2016rl,wang2016learning} and have recently been combined with RL to solve nonstationary problems via model-free~\citep{finn2017model,houthooft2018evolved,rothfuss2018promp} and model-based approaches~\citep{clavera2018learning, smundsson2018meta}. 
The closest work to our research is \citep{nagabandi2018deep}, which uses a mixture of DNNs as the dynamics model in model-based RL but still requires a model-agnostic meta-learning (MAML) prior. 
Researchers in~\citep{vuorio2019multimodal} augment MAML with a multimodal task distribution for solving substantially different tasks. However, it inherits the limitations of MAML (such as assuming accessible task simulations and clear task boundaries). 
Additionally, the meta training strategy increases sample complexity and thus makes many model-free meta-learning algorithms infeasible to implement on real-world problems~\citep{peters2008reinforcement}.
Using a single GP as the dynamics model in model-based meta-RL is explored in~\citep{smundsson2018meta} by updating latent variables for different tasks but in an offline and episodic setting. 
To the best of our knowledge, our method is the first capable of robustly handling online nonstationary tasks without requiring task delineations or depending on a pre-trained model.

Using a mixture of experts to solve different tasks is explored in continual learning~\citep{wilson2007multi,chen2018lifelong, candy1991self} and Multi-task Learning~\citep{ruder2017overview, evgeniou2004regularized, jacob2009clustered}. However, previous works from both communities require clear task delineations.
The conventional inference methods for mixture models mainly consist of MCMC sampling~\citep{jain2004split, dahl2005sequentially} and variational inference~\citep{blei2006variational}. 
Both methods keep the information of the whole dataset and do inference iteratively via multiple passes. 
In contrast, 
% In many real-world situations, data acquired by the agent keeps increasing with continued interaction with the environment. 
% Therefore, we seek to 
streaming variational inference~\citep{broderick2013streaming, huynh2016streaming, tank2015streaming} for Bayesian nonparametric models is designed for handling streaming data and requires a single computation pass. 
Sequential variational inference~\citep{lin2013online} for DP mixtures has been recently integrated with DNNs for image classification~\citep{Lee2020A}.

%================================ Preliminaries ================================

% \section{Model-based RL with Infinite Mixtures of Gaussian Processes Model Overview}
\section{Model-Based RL with an Infinite Mixture of Gaussian Processes}

\label{sec:modeloverview}

% In this section, we present the two modules in the proposed framework: the prediction module, namely the learned system dynamics model as the infinite mixtures over function space, and the planning module which is based on Model Predictive Control. 

% \subsection{Infinite Mixtures of Gaussian Processes as dynamics models}
% \label{sec:stochasticprocesses}
Model-based RL algorithms rely on the learned dynamics model to roll out environments. 
For real-world tasks that contain substantially different dynamics, using an infinite mixture model as the dynamics model alleviates the harmful mutual knowledge transfer when using a single model, and enables the backward transfer of knowledge by recalling and updating stored dynamics models. 
Learning the system dynamics model $f$ is carried out by performing inference on data-efficient GPs to avoid training a prior model as in~\citep{jerfel2019reconciling, nagabandi2018deep}. Additionally, GPs define distributions over functions and thus naturally capture the aleatoric uncertainty of noisy environments. 
We use Model Predictive Control (MPC) for selecting an action at each time step, which can be seen as a closed-loop controller and increases robustness to model errors.

We consider the system dynamics model in the from of $\xv_{t+1} =\xv_t + f(\xv_t, \uv_t)$. The input is augmented as $\txv_t = (\xv_t,\uv_t)$, where $\xv_t \in \mathbb{R}^c$ and $\uv_t \in \mathbb{R}^d$ are the state and action at time $t$, respectively. The target $\yv_t = \Delta \xv_t= \xv_{t+1} - \xv_t$ is the state increment. 
A history dataset $\mathcal{D}= \{\txv^j, \yv^j \}_{j=1}^{m}$ is maintained to update the mixture model and predict increments in MPC. The inputs and targets are aggregated into $\tBigX \in \mathbb{R}^{(c+d) \times m }$ and $Y \in \mathbb{R}^{c \times m}$.
% \textbf{Gaussian Processes.}    
With each dynamics model as a GP~\citep{rasmussen2003gaussian}, the dynamics function $f$ can be decoupled by dimension as $f = (f_1, ..., f_c)$ with $f_i: \mathbb{R}^c \rightarrow \mathbb{R}$.   The state difference given a new observation $\xv$ and action $\uv$ is drawn from the predictive distribution 
\begin{align}
    p(f | \mathcal{D}, \txv_{t}) =& \prod_{i=1}^{c} \mathcal{N}(\text{m}(f_i), \text{cov}(f_i)). \label{eq:GP}
\end{align}
The mean function is $ \text{m}(f_i) =K_i(\txv, \tBigX)[K_i(\tBigX, \tBigX) + \sigma_{i}^2 I]^{-1}Y^i$ and the covariance matrix is $\text{cov}(f_i) = K_i(\txv, \txv)  - K_i(\txv, \tBigX)[K_i(\tBigX, \tBigX) + \sigma_{i}^2 I]^{-1} K_i(\tBigX, \txv)$. 
$Y^i$ denotes the target's $i$th dimension. 
$\sigma_{i}$ is the standard deviation of observation noise of dimension $i$.
The matrix $K_i$ is fully specified by the kernel function $k_i(\txv, \txv_{*})$, which defines the function smoothness. For simplicity, we use the scaled squared exponential kernel $k_i(\txv, \txv_{*}) = w_{i}^2 \exp ( -\frac{1}{2} \sum_{j=1}^{c+d} w_{i,j}(\txv^j- \txv_{*}^j)^2 )$.
% \begin{equation}
%     k_i(\txv, \txv_{*}) = \sigma_{ARD,i}^2 \exp ( -\frac{1}{2} \sum_{j=1}^{c+d} w_{i,j}(\txv^j- \txv_{*}^j) ). \label{eq:kernel}
% \end{equation}
$w_{i,j}$ is the reciprocal of the lengthscale between dimensions $i$ and $j$, and $w_i$ is the output scale of dimension $i$.

\section{Scalable Online Bayesian Inference}
\label{sec:methods}

% present key assumptions
% a brief overview of the general structure of algorithms
% point to corresponding section 

% model selection, for prediction and training the 
This section presents a scalable online Bayesian inference method for learning an infinite mixture of GPs in the model-based RL setting. 
% With the episodic assumption released, w
We assume that it is possible for a new type of dynamics to emerge at every timestep, and the total number of different dynamics is unknown. 
We observe a new data pair $(\xv_t, \uv_t, \Delta\xv_t)$ each time $t$ the RL agent interacts with the environment. 
To learn the mixture model, we first identify the latent dynamics assignment variable $z_t$ and then update the dynamics-specific parameters $\thetav_{z_t}$ of expert $M_{z_t}$.
Our goal is to jointly do inference over $z$ and learn $\thetav$ in an online streaming setting by maximizing the log posterior probability given the observed data $p_n(z_{1:n}, \thetav | \mathcal{D})$. 
To get a reasonable action at time $t+1$, with the inferred dynamics assignment $z_t$, we select the expert $M_{z_t}$ to generate predictions $\yv_{t+1:t+T}$ for the MPC. 

We first introduce sequential variational inference with transition prior in Section~\ref{sec:svi-tp}. 
To make our model scalable to large datasets, we introduce an optimization-based method in Section~\ref{sec:datadistillation} to eliminate redundant data points.
% in order to decrease the spatial and computational complexity during prediction. 
The stability of the mixture model is enhanced by merging similar experts and pruning redundant ones, as in Section~\ref{sec:mergeprune}. We present the overall pipeline in Algorithm~\ref{algo:pipeline} and discuss the effect of model parameters in Section~\ref{section:suppl:parameters} in the supplementary material.

\begin{algorithm}[t]
\SetAlgoLined
\SetNoFillComment
% \setstretch{1.12}
\KwIn{Concentration parameter $\alpha$, Initial parameter $\thetav_0$, Sticky parameter $\beta$, KL threshold $\epsilon$, Merge trigger $n_{merge}$, Data Distillation trigger $n_{distill}$, Inducing point number $m$ }
\KwOut{Infinite Mixture Model $\Mv$, Representative Dataset $\mathcal{D}$}
    Initialization: $\Mv \leftarrow \{ M_0 \}$, $\mathcal{D} \leftarrow  \{ \emptyset \}$, $t\leftarrow  0$, $z_0 \leftarrow 0$, and $K \leftarrow 1$\;
    $(\xv_0,\uv_0,\Delta \xv_0) \leftarrow Random Policy$\;
    $\mathcal{D} \leftarrow \mathcal{D} \cup \{(z_0, \xv_0,\uv_0,\Delta \xv_0)\}$ \;
    Update $\thetav_0$ with (\ref{eq:theta_update})\;
    \While{task not finish}{
    $z_{old} \leftarrow z_t$, $t \leftarrow t+1$\;
    $(\xv_t,\uv_t,\Delta \xv_t)\leftarrow MPC(M_{z_{t-1}})$\;
    \tcp{Sequential Variational Inference with Transition Prior (Section~\ref{sec:svi-tp})}
    Update $q_t^{pr}(z_t)$ with (\ref{eq:transitionprior})\;
    Update $z_t = \text{argmax}_{k} \rho_t(z_{tk})$ with (\ref{eq:softassignment})\;
    \If{$z_t = K$}{ Append $M_{z_t}$ to $\Mv$, $K\leftarrow K+1$}
    $\mathcal{D} \leftarrow \mathcal{D} \cup \{(z_t, \xv_t,\uv_t,\Delta \xv_t)\}$\;
    Update $\thetav_{z_t}$ with (\ref{eq:theta_update})\;
    
    \tcp{Expert Merge and Prune (Section~\ref{sec:mergeprune})}
    \If{ $ \sum_{i=0}^t \mathbf{1}\{z_i = z_t\} = n_{merge}$  }{
    $d_{t}(k) =  d(M_{z_t}, M_k), \ k=0,...,K-1$ with (\ref{eq:merge})\;
    \If{ $\min_k d_{t} \leq \epsilon$ }{ 
    Merge $M_{z_t}$ to the most similar model $M_{\arg \min_k d_t}$, $K \leftarrow K-1$}
    }
    \If{ $ \sum_{i=0}^t \mathbf{1}\{z_i = z_{old}\} \leq n_{merge}$ {\bf and} $z_t \not = z_{old}$}{
    Merge $M_{z_t}$ to the most similar adjacent model based on $d(M_{z_t}, M_k)$, $K \leftarrow K-1$}
    
    \tcp{Data Distillation (Section~\ref{sec:datadistillation})}
    \If{ $ \sum_{i=0}^t \mathbf{1}\{z_i = k\} \geq n_{distill}, \ \forall k$ }{
    Get $m$ inducing points with (\ref{eq:elboinducing})}    
    }
    \caption{Bayesian Inference for Continual Online Model-based Reinforcement Learning}
    \label{algo:pipeline}
\end{algorithm}

\subsection{Sequential Variational Inference with Transition Prior}
\label{sec:svi-tp}

% In streaming variational inference, the posterior distribution $p_n(z_{1:n}, \thetav | \mathcal{D})$ can be written as product of factors. 
We approximate the intractable $p_n(z_{0:n}, \thetav | \mathcal{D})$ as $\hat{q}_n(z_{0:n}, \thetav) = \prod_{k=0}^{\infty} \gamma_n(\thetav_k) \prod_{i=0}^{n} \rho_n(z_i)$ using Assumed Density Filtering (ADF) and mean field approximation.
As derived in~\citep{tank2015streaming}, the optimal distribution of the dynamics assignment $z_n$ of the $n$th observation is
\begin{align}
    \rho_n(z_{nk}) \propto  
    \begin{cases}
    q^{pr} (z_{nk}) \int p( (\txv_n, \yv_n) | \thetav_{k}) \gamma_{n-1}(\thetav_k) d \thetav_k  & 0 \leq k \leq K_{n-1}-1 \\
    q^{pr} (z_{nk}) \int p( (\txv_n, \yv_n) | \thetav_k) G_0(\thetav_k) d \thetav_k  & k = K_{n-1} 
    \end{cases}\label{eq:softassignment}
\end{align}
where $q^{pr}(z_n) = \sum_{z_{0:n-1}} p(z_n | z_{0:n-1}) \prod_{i=0}^{n-1} \rho_{n-1}(z_i)$, and $G_0$ is the base distribution for dynamics-specific parameter $\thetav$. $q^{pr}$ acts as the prior probability of the assignments and is analogous to the predictive rule $p(z_n | z_{0:n-1})$. 
% The integration term is the likelihood conditioned on different dynamics models.
$ G_0$ and $\gamma_{n-1}$ in general are in the same exponential family as $\rho_n(z_{n})$ so that (\ref{eq:softassignment}) is in closed form.
When using a DP mixture model, $q^{pr}$ is in the form of the Chinese Restaurant Process prior. The sequential variational inference method for DP mixtures~\citep{lin2013online} is suitable for dealing with streaming data with overall complexity $O(NK)$, where $N$ is the number of data points and $K$ is the expected number of experts.
% as in Equation~\ref{eq:CRP}, where $\alpha$ is the concentration parameter.
% \begin{align}
%     q^{pr} \propto \begin{cases} 
%     \sum_{i=1}^{n-1} \rho_i(z_{ik}), & k \leq K_{n-1} \\
%     \alpha, & k = K_{n-1} +1 
%     \end{cases}
%     \label{eq:CRP}
% \end{align}

However, in the task-agnostic setting where the task assignment is evaluated at each time step, dynamics models are not independently sampled from a prior. Instead, the adjacent observations tend to belong to the same dynamics model, and there may exist temporal transition dependencies between different dynamics models.
% different dynamics models are likely to hold temporal dependency. 
Therefore, we adopt the Markovian assumption when selecting the assignment prior $q^{pr}_{n}$ and propose a transition prior that conditions on the previous data point's dynamics assignment $z_{n-1}$ as follows:
\begin{align}
    q_n^{pr} \propto \begin{cases} 
    \sum_{i=1}^n \mathbf{1}\{z_{i-1} = z_{n-1}\} \rho_i(z_{ik}) + \mathbf{1}\{k = z_{n-1}\}\beta,  & 0 \leq k \leq K_{n-1}-1 \\
    \alpha, & k = K_{n-1} 
    \end{cases} \label{eq:transitionprior}
\end{align}
Here $\beta$ is the sticky parameter that increases the probability of self-transition to avoid rapidly switching between dynamics models~\citep{fox2011sticky}. 
Note that in Algorithm~\ref{algo:pipeline}, we approximate the soft dynamics assignment $\rho_i(z_{ik})$ of the $i$th data pair with the hard assignment $z_i = \text{argmax}_{k} \rho_i(z_{ik})$. The argmax/hard assignment approximation splits the streaming data and thus accelerates the stochastic update of model parameters and simplifies the form of transition prior. However, this involves a trade-off \citep{kearns1998information} between how well the data are balanced and the density accuracy.

% In this paper, we use $\beta=1$.

For GPs, $\thetav$ is a set of kernel parameters $\{ w_{i}, w_{i,1}, ... , w_{i,c}, \sigma_{i} \}_{i=1}^{c}$.
Since it is hard to find a conjugate prior $G_0$ for $\thetav$, and we are considering a task-agnostic setting with limited prior knowledge, we use the likelihood $F((\txv_n, \yv_n) | \thetav_{k} ) $ to approximate the integrals in (\ref{eq:softassignment}), inspired by the auxiliary variable approach \citep{neal2000markov}.  
% { \color{blue} \sout{we hope to decrease the effect of the prior $G_0$ and enlarge that of the collected data.} }  
In order to create a GP expert for unobserved dynamics, we need to find the likelihood of a GP with no data \citep{rasmussen2002infinite}.
Therefore, we initialize the kernel of the first GP dynamics model $M_0$ with a fixed reasonable point estimation $\thetav_{init}$ to evaluate the likelihood conditional on it.
We then train a global GP dynamics model with all online collected data and use the updated parameters as the prior when initializing succeeding GP experts.
To deal with the non-conjugate situation, we use stochastic batch optimization and update $\thetav$ with (\ref{eq:theta_update}).
% Equation~\ref{eq:theta_update}. 
\begin{equation}
    \thetav_k \leftarrow \thetav_k - \eta \sum_{i=0}^n \mathbf{1}\{z_i = k\}  \nabla_{\thetav} \log p(\yv_i | \txv_i, \mathcal{D}, \thetav_k), \ \forall k = 0,..., K_n -1
    \label{eq:theta_update}
\end{equation}

% ============ Involve Neural Processes
% For Gaussian Processes and Neural Procsses, $\thetav$ is a set of kernel parameters $\{ w_{i}, w_{i,1}, ... , w_{i,c}, \sigma_{i} \}_{i=1}^{c}$ and neural network weights, respectively. In both cases, it is hard to find a conjugate prior of the function approximator. 
% For simplicity, we initialize the kernel of the first Gaussian Process dynamics model $M_0$ with a fixed reasonable point estimation $\thetav_{init}$ and train a global Gaussian Process dynamics model with all collected data and use its kernel parameters as the prior for newly initialized ones.
% For simplicity, we initialize the kernel of new Gaussian Process dynamics models with
% a fixed reasonable point estimation $\thetav_{init}$. 
% \textcolor{red}{TODO: double check NP initialization: The weights of a new Neural Process dynamics model are initialized with Model-Agnostic Meta Learning as in~\citep{jerfel2019reconciling, nagabandi2018deep}.}

Note that the iterative updating procedure of dynamics assignment $z$ and dynamics-specific parameters $\thetav$ can be formulated as a variational expectation-maximization (EM) algorithm. In the variational E step, we calculate the soft assignment, which is the posterior probability conditioned on the dynamics parameters $\thetav$. In the M step, given each data pair's assignment, we update $\thetav$ for each dynamics model with the conjugate gradient method to maximize the log-likelihood.

\subsection{Data Distillation with Inducing Points}
\label{sec:datadistillation}

GP dynamics models need to retain a dataset $\mathcal{D}$ for training and evaluation, for which the spatial and computational complexity increases as more data is accumulated. 
For instance, the prediction complexity in (\ref{eq:GP}) is $O(N_k^3)$ due to the matrix inversion. 
To make the algorithm computationally tractable, we select a non-growing but large enough number of data points that contain maximum information to balance the tradeoff between algorithm complexity and model accuracy. 
In a model-based RL setting, the data distillation procedure is rather critical since the agent exploits the learned dynamics models frequently to make predictions in MPC, and thus the overall computational load heavily depends on the prediction complexity.

For each constructed dynamics model $k$, data distillation is triggered if the number of data points belonging to $M_k$ reaches a preset threshold $n_{distill}$. 
Define $\mathcal{D}_k = \{(z_i, \xv_i, \uv_i, \Delta \xv_i) \ |  \ z_i=k \}$.
The $m$ inducing points $\mathcal{D}_{k,m}$ are selected while preserving the posterior distribution over $f$ as in exact GP in (\ref{eq:GP}) by maximizing the following lower bound that is widely used in Sparse GP literature~\citep{titsias2009variational, tran2015variational}:
\begin{align}
    L = \sum_{i=1}^c \Big[ \log  \mathcal{N} (Y^i;0, K_{nm}K_{mm}^{-1}K_{mn} +I\sigma_i^2)
    - \frac{1}{2\sigma_i^2} \Tr (K_{nn} - K_{nm}K_{mm}^{-1}K_{mn}) \Big]
\label{eq:elboinducing}
\end{align}
% Maximizing the objective $L$ is equivalent to minimizing the KL-divergence between $q(f,f_m)$ and $p(f,f_m |Y)$, where $f_m$ are function values evaluated at $\mathcal{D}_{k,m}$. 
Note that the collected data depends on the learned dynamics model and the rolled-out policy, which introduces a nonstationary data distribution. 
In addition to the setting that the GP experts are trained online from scratch, treating inducing points as pseudo-points and optimizing (\ref{eq:elboinducing}) via continuous optimization may lead to an unstable performance in the initial period.  
Instead, we directly construct $\mathcal{D}_{k,m}$ by selecting real data points from $\mathcal{D}_{k}$ via Monte Carlo methods. 
Other criteria and optimization methods for selecting inducing points can be found in~\citep{titsias2019functional}.

\subsection{Expert Merge and Prune}
\label{sec:mergeprune}

Since each data point's dynamics assignment is only evaluated once, dynamics models may be incorrectly established, especially in early stages where the data numbers of existing dynamics models are relatively small.
Redundant dynamics models not only harm the prediction performance by blocking the forward and backward knowledge transfer but also increase the computational complexity when calculating $z$ (which requires iterating over the mixture).
Therefore, we propose to merge redundant models and prune unstable ones with linear complexity w.r.t. the number of spawned dynamics models. 
Other methods for merging experts can be found in~\citep{guha2019posterior}.
% . The proposed merge and prune mechanism is designed to be suitable for online settings, as the activation number scales 

\textbf{Expert merging mechanism.}     
We say that a newly constructed dynamics model $M_K$ is in a burn-in stage if the number of data points belonging to it is less than a small preset threshold $n_{merge}$.
At the end of the burn-in stage, we check the distances between $M_K$ with the older models.
If the minimum distance obtained with $M_{k'}$ is less than the threshold $\epsilon$, we merge the new model $M_K$ by changing the assignments of data in $\mathcal{D}_K$ to $z=k'$. We use the KL-divergence between the predictive distributions evaluated at $\mathcal{D}_K$ conditioned on two dynamics models as the distance metric:
\begin{equation}
    d(M_K, M_k) = \sum_{i \in \mathcal{D}_K}  KL (p( f | \mathcal{D}_k, \txv_i ) \| p(f | \mathcal{D}_K, \txv_i) ) 
    \label{eq:merge}
\end{equation}

\textbf{Expert pruning method.}    
We assume that in real-world situations, each type of dynamics lasts for a reasonable period.
If a dynamics model $M_K$ accumulates data less than the merge trigger $n_{merge}$ when a new dynamics model is spawned, we treat $M_K$ as an unstable cluster and merge it with adjacent dynamics models based on the distance in (\ref{eq:merge}).

%================================ Experiments ================================

\section{Experiments: Task-Agnostic Online Model-Based RL}
\label{sec:experiment}

We present experiments in this section to investigate whether the proposed method (i) detects the change points of the dynamics given the streaming data, (ii) improves task detection performance by using the transition prior, (iii) automatically initializes new models for never-before-seen dynamics and identifies previously seen dynamics, and (iv) achieves equivalent or better task performance than baseline models while using only a fraction of their required data.

\begin{wrapfigure}{r}{0.5\textwidth}
\centering
\includegraphics[width=6.5cm]{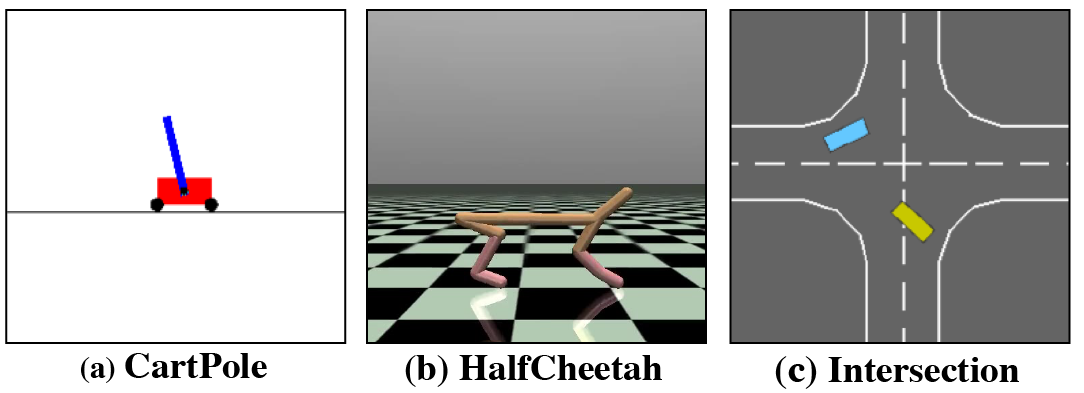}
\caption{Simulation environments}
\label{fig:envs}
\end{wrapfigure}

We use three non-stationary environments for our experiments, as shown in Figure~\ref{fig:envs}. 
In each environment, the RL agent cyclically encounters several different types of dynamics.
In $\mathtt{Cartpole}$-$\mathtt{Swing Up}$, we aim to swing the pole upright meanwhile keep the cart in the middle. The pole length $l$ and mass $m$ vary sequentially by alternating between four different combinations.
$\mathtt{HalfCheetah}$ is a more complex control problem having larger state and control dimensions with the non-stationary dynamics induced by different torso loads. 
The goal is to make the half cheetah run as fast as possible.
$\mathtt{Highway}$-$\mathtt{Intersection}$ contains surrounding vehicles with different target destinations, and the ego vehicle needs to avoid collisions while heading to its goal.
More detailed experiment settings are presented in Section~\ref{section:suppl:simulations} in the supplementary material.

We compare our method with five model-based baselines detailed as follows: 
% including three streamlined versions of our method that adapt a pretraining and online adaptation setting, and two meta-learning based approaches:
\begin{enumerate}
\item[(a)] \textbf{Single dynamics model}: 
We investigate the task performance of using a GP, a DNN, and an Attentative Neural Process (ANP)~\citep{kim2018attentive, qin2019recurrent} as the dynamics model.
The GP is trained online from scratch, while the ANP and the DNN are first trained offline with data pre-collected in all dynamics that the agent may encounter online. All three models are updated online to account for the non-stationarity of environments after each batch of data is collected. 
% The Gaussian Process is updated at each time step while the other two are updated after collecting a batch of data.

% Since just a single GP is used, the non-stationary nature of environments is ignored.

\item[(b)] \textbf{Mixture of DNNs}: We compare with another two pre-trained baselines using mixtures of DNNs as the dynamics model with DP prior (DPDNN) and the proposed transition prior (T-DNN), respectively. Considering that our method does not require an adaptation module for updating model parameters, we directly use the weights of a global DNN that is trained with all collected data as the prior instead of using a MAML-trained prior as in~\citep{nagabandi2018deep}. 
\item[(c)] \textbf{Meta-RL}: We use the MAML algorithm~\citep{finn2017model} to learn the dynamics model for model-based RL. After being pre-trained with several episodes, the meta-model is adapted online with recently collected data to deal with nonstationarity.
\end{enumerate}

\subsection{Task Switching Detection}
\label{Exp:SwitchingDetection}

\begin{figure}[t]
    \centering
    \includegraphics[width=11.0cm]{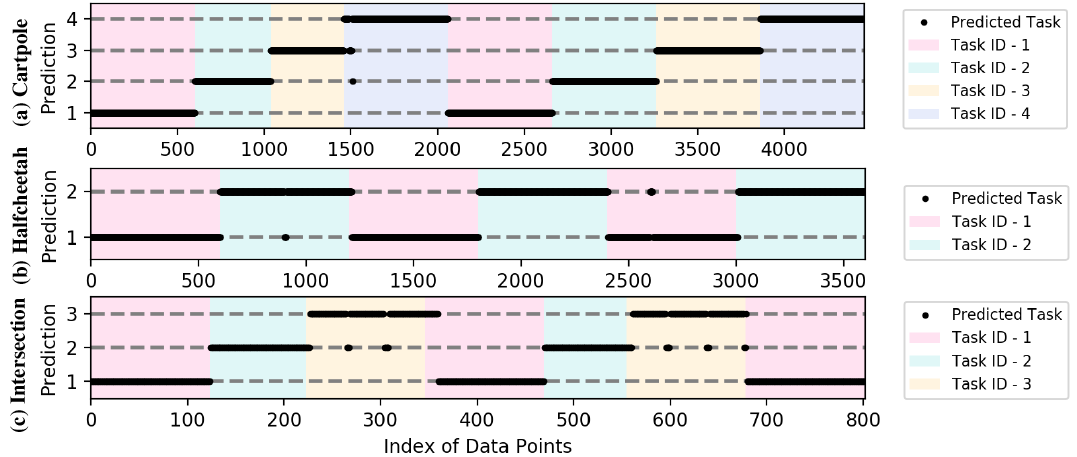}
    \caption{Dynamics assignments with the proposed transition prior. 
    % The environments from top to bottom are (a) $\mathtt{Cartpole}$-$\mathtt{Swing Up}$ (b) $\mathtt{HalfCheetah}$ (c) $\mathtt{Intersection}$. 
    Our method successfully detects the dynamics shift. It allocates new components to model previously unseen types of dynamics and recalls stored models when encountering seen dynamics.}
    \label{fig:cluster_cp_transition}
\end{figure}

\begin{figure}[t]
    \centering
    \includegraphics[width=\textwidth]{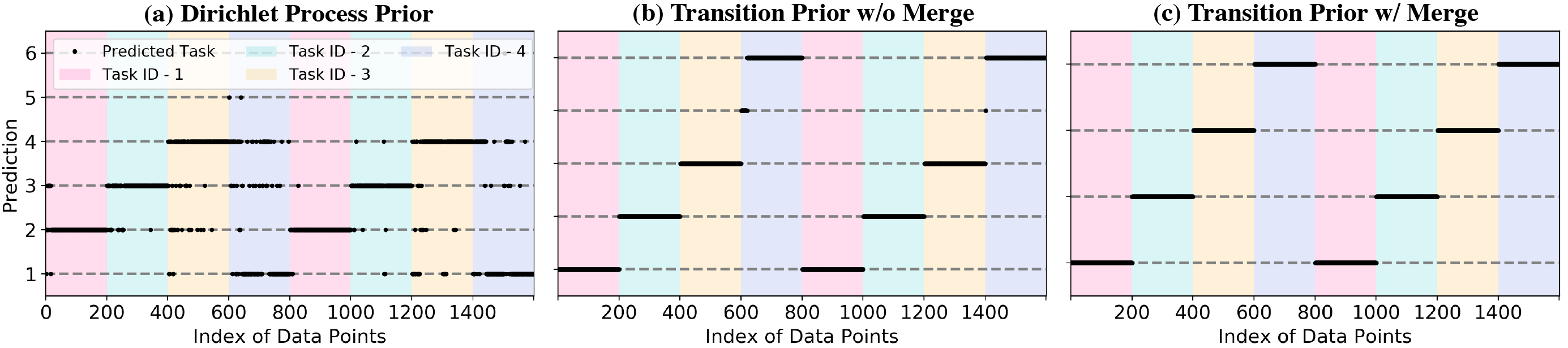}
    \caption{Ablation experiment results of the transition prior as well as the merge and prune mechanism in $\mathtt{Cartpole}$-$\mathtt{Swing Up}$. The proposed transition prior achieves more accurate dynamics assignments than the DP prior. The merge and prune mechanism successfully merges redundant dynamics models.}
    \label{fig:cluster_cp_DP}
\end{figure}

In all three environments that contain sudden switches between different dynamics, our method detects the task distribution shift as visualized in Figure~\ref{fig:cluster_cp_transition}. 
In general, the predicted region of each type of dynamics matches that of the ground truth. 
We notice that there exists some delay when detecting change points and hypothesize that this may be due to the clustering property and sticky mechanism of the transition prior.
% With the \textcolor{red}{normalized Dirichlet Process prior}, our method keeps initializing new dynamics models when the dynamics change and fails to recall previously stored models, as can be seen in Fig.\ref{fig:cluster_cp_DP}(a).
Although DPs~\citep{teh2010dirichlet} have the so-called rich-gets-richer property, directly using a DP prior fails to capture the temporal relationship between data points and thus leads to inaccurate dynamics assignments (Figure~\ref{fig:cluster_cp_DP}~(a)).
We also notice that wrong dynamics assignments result in unstable task performances with smaller rewards and larger variances, as shown in Section~\ref{section:suppl:expresults}. 
There are some redundant dynamics models during online training (Figure~\ref{fig:cluster_cp_DP}~(b)) when not using the merge and prune mechanism. When comparing Figure~\ref{fig:cluster_cp_DP}~(b) and~(c), we can see that our method successfully merges redundant dynamics models to the correct existing ones.

\subsection{Task Performance}
\label{Exp:taskreward}

\begin{figure}[t!]
    \centering
    \includegraphics[width=0.95\textwidth]{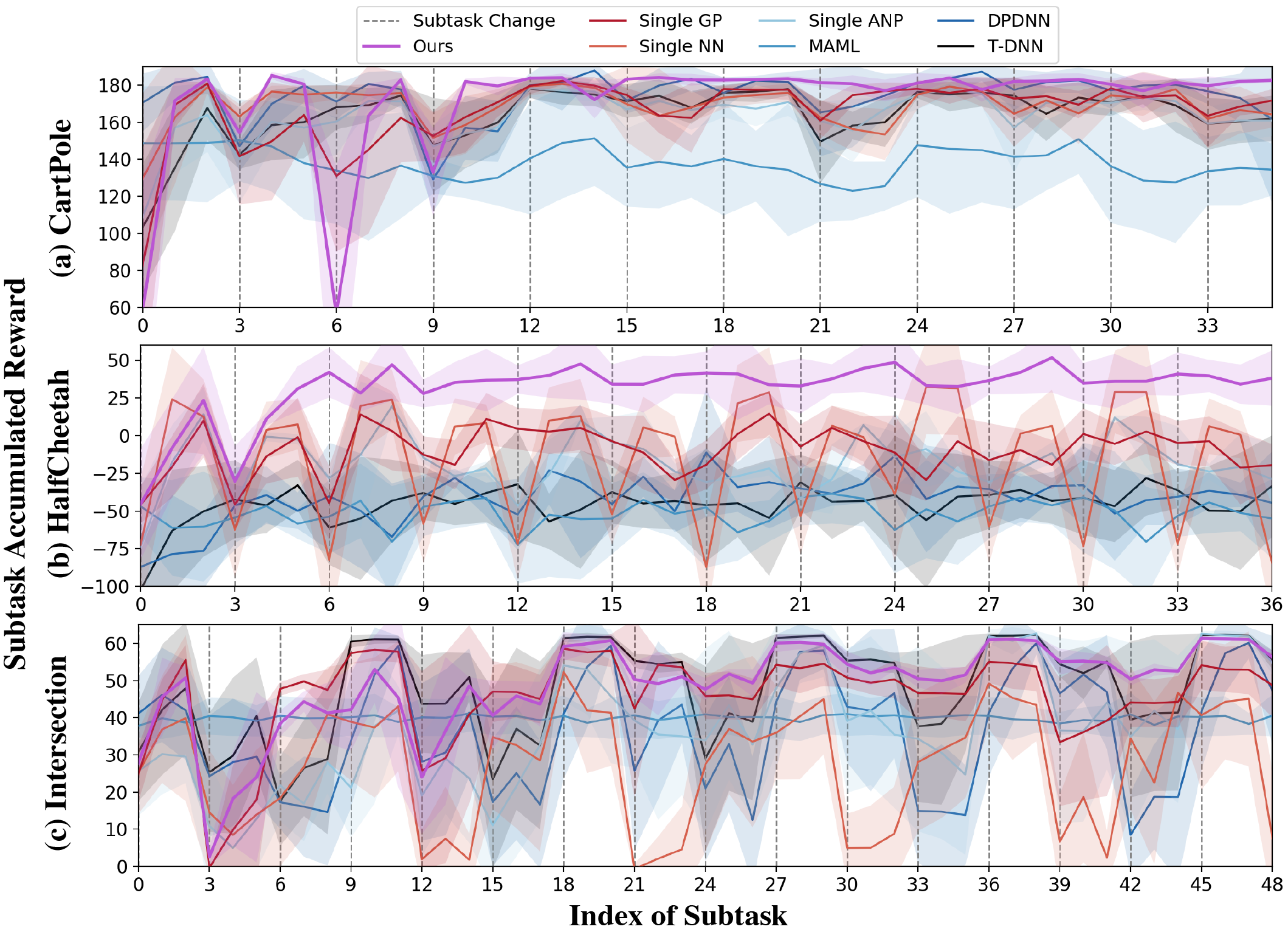}
    \caption{Accumulated rewards in three nonstationary environments. A subtask is analogous to an episode. Each vertical dotted line indicates a dynamics switch. 
    % Each baseline in each environment is run with 10 random seeds. 
    Our method is more robust, data-efficient, and has higher rewards.
    % All the baselines except for GP is pre-trained with 
    }
    \label{fig:cartpole-subtask-rewards}
\end{figure}

We evaluate the task performance in terms of the accumulated rewards for each type of dynamics, as displayed in Figure~\ref{fig:cartpole-subtask-rewards}. 
Since a separate model is initialized and trained from scratch whenever a new type of dynamics is detected, our method's reward oscillates during the initial period after the dynamics shift (especially in Figure~\ref{fig:cartpole-subtask-rewards}~(a)). 
However, our method performs well in a new type of dynamics after collecting just a handful of data points that are far less than DNN-based methods. For example, for each type of dynamics in $\mathtt{Cartpole}$-$\mathtt{Swing Up}$, our method converges after collecting 600 data points, while DNN may need around 1500 data points~\citep{chua2018deep}.
% However, our method performs well in a new type of dynamics after collecting hundreds of data point, which is far less than the amount of data required by methods using DNNs to achieve similar performance.
By learning the latent dynamics assignment, our method quickly adapts when it encounters previously seen dynamics by shifting to the corresponding dynamics model.
These previously learned models are further improved by being updated with the newly collected data according to their dynamics assignments. 
Additionally, our method is more robust than all the baselines and has a smaller variance at convergence.  

Using a single dynamics model automatically preserves global dynamics properties. Therefore using a single GP without pre-training does not suffer apparent performance degradation when the dynamics switch. 
Since both the DNN and the ANP use pre-trained models, they achieve equivalent (in complex $\mathtt{Highway}$-$\mathtt{Intersection}$) or better (in simple $\mathtt{Cartpole}$-$\mathtt{Swing Up}$) performance as our method when encountering new dynamics.
However, the three single dynamics model baselines fail to consistently succeed and oscillate a lot as the environment getting more complex. 
% because they lack the ability to separate different types of dynamics.
For example, when using a DNN in the unprotected left-turn scenario in $\mathtt{Highway}$-$\mathtt{Intersection}$, the RL agent keeps colliding with the other vehicle.
Although the ANP is a DNN approximation of the GP and can deal with multiple tasks itself~\citep{galashov2019meta}, its performance still oscillates, and thus it cannot robustly handle the multimodal task distribution. 
The online adaptation mechanism causes the single model to tend to overfit to the current dynamics. 
The overfitting problem further harms the quick generalization performance since the adjacent dynamics are substantially different, and previously adapted model parameters may be far from the optimal ones for the new dynamics. 
Additionally, without task recognition, the model suffers from the forgetting problem due to not retaining data and parameters of previously seen dynamics.

Figure~\ref{fig:cartpole-subtask-rewards} also show that our method outperforms the three baselines that aim to handle nonstationarity. 
The model-based RL with MAML is easily trapped in local minima determined by the meta-prior. For instance, the RL agent in $\mathtt{Highway}$-$\mathtt{Intersection}$ learns to turn in the right direction but fails to drive to the right lane. 
In our experiments, MAML also cannot quickly adapt to substantially different dynamics.
This is because MAML suffers when a unimodal task distribution is not sufficient to represent encountered tasks. 
The model-based RL with a DP mixture of DNNs performs slightly better than a single DNN but still oscillates due to inaccurate task assignments, as in Section~\ref{section:suppl:expresults}. T-DNN performs similarly to DPDNN and still underperforms our method due to the inaccuracy of DNN model predictions with limited data.

% Since both the DNN and the ANP use pre-trained models to initialize new clusters, they achieve equivalent (in complex $\mathtt{Highway}$-$\mathtt{Intersection}$) or better (in simple $\mathtt{Cartpole}$-$\mathtt{Swing Up}$) performance when new dynamics are first encountered.

%================================ Conclusion ================================

\section{Discussion}
\label{sec:discussion}
% conclusion, limitation and future work

We propose a scalable online model-based RL method with an infinite mixture of GPs to handle real-world task-agnostic situations with limited prior knowledge. 
Our method achieves quick adaptation and avoids pre-training by using data-efficient GPs as dynamics models and avoids catastrophic forgetting by retaining a mixture of experts. 
Our model performs well when dynamics are substantially different by constructing a multimodal predictive distribution and blocking harmful knowledge transfer via task recognition. 
We propose a transition prior to explicitly model the temporal dependency and thus release the assumption that tasks are independent and identically distributed. 
Additionally, our method detects the dynamics shift at each time step, so it is suitable for situations with unknown task delineations. 
We learn the mixture via online sequential variational inference that is scalable to extensive streaming data with data distillation and the merge and prune technique. 
Since computing the posterior of a GP becomes intractable as the data size and data dimension increase, replacing GPs with Neural Processes~\citep{kim2018attentive, pmlr-v80-garnelo18a} would be an interesting direction to explore.
Another direction would be to incorporate meta-learning into our method to better leverage the commonly shared information across the different dynamics.

% \clearpage

\section*{Broader Impact}

% positive: 
% robustness to different tasks, 
% generalization to other online clustering tasks (beyond RL framework), Interpretability due to the task specification,

% what might be the possible applications and their positive and negative impacts  

% negative: 
% over-trust, it is possible that task assignments are wrongly assigned.  
% Interpretability -> misuse of outputs, 

% In this paper, researchers propose a model-based reinforcement learning method with an infinite mixture of Gaussian Process as the dynamics model for solving online nonstationary tasks.

This work is a step toward General Artificial Intelligence by eliminating the pre-train stage and equipping reinforcement learning agents with the quick-adaptation ability.
The proposed method could be applied in a wide range of applications when the prior knowledge is not accessible or not beneficial, including space rover navigation, rescue robot exploration, autonomous vehicle decision making, and human-robot interaction.

Our research increases algorithmic interpretability and transparency for decision-making by providing inferred task assignments. 
It also enhances the algorithm's robustness by explicitly separating different types of tasks and thus increases users' trust.
Additionally, our research improves algorithmic fairness by not relying on prior knowledge that may contain human-induced or data-induced biases. 
At the same time, our research may have negative impacts if misused. The potential risks are listed as follows: (1) Over-trust in the results (e.g., the task assignments) may lead to undesirable outcomes. (2) If used for illegal purposes, the model may instead enlarge the possible negative outcome due to its explainability. (3) The task assignment results may be misinterpreted by those who do not have enough related background. (4) The evolving online nature of the method may increase the uncertainty and thus mistrust of human collaborators. Note that our method inherits the possible positive and negative impacts of reinforcement learning, not emphasized here.

To mitigate the possible risks mentioned above, we encourage research to 
(1) investigate and modulate the negative impacts of the inaccurate information provided by algorithms, 
(2) understand how humans interact with evolving intelligent agents in terms of trust, productivity, comfort level, 
(3) find out the impact of using our model in real-world tasks.

\begin{ack}

% \section*{Acknowledgements and Disclosure of Funding}
Thanks go to Mr. Keegan Harris for revision and advice. The authors gratefully acknowledge the support from the Manufacturing Futures Initiative at Carnegie Mellon University, made possible by the Richard King Mellon Foundation. The ideas presented are solely those of the authors.
\end{ack}

\bibliographystyle{unsrtnat}
\bibliography{main}

\begin{thebibliography}{50}
\providecommand{\natexlab}[1]{#1}
\providecommand{\url}[1]{\texttt{#1}}
\expandafter\ifx\csname urlstyle\endcsname\relax
  \providecommand{\doi}[1]{doi: #1}\else
  \providecommand{\doi}{doi: \begingroup \urlstyle{rm}\Url}\fi

\bibitem[Lake et~al.(2015)Lake, Salakhutdinov, and Tenenbaum]{lake2015human}
Brenden~M Lake, Ruslan Salakhutdinov, and Joshua~B Tenenbaum.
\newblock Human-level concept learning through probabilistic program induction.
\newblock \emph{Science}, 350\penalty0 (6266):\penalty0 1332--1338, 2015.

\bibitem[Sæmundsson et~al.(2018)Sæmundsson, Hofmann, and
  Deisenroth]{smundsson2018meta}
Steindór Sæmundsson, Katja Hofmann, and Marc~Peter Deisenroth.
\newblock Meta reinforcement learning with latent variable gaussian processes.
\newblock In \emph{Conference on Uncertainty in Artificial Intelligence (UAI)},
  May 2018.
\newblock URL
  \url{https://www.microsoft.com/en-us/research/publication/meta-reinforcement-learning-latent-variable-gaussian-processes/}.

\bibitem[Clavera et~al.(2019)Clavera, Nagabandi, Liu, Fearing, Abbeel, Levine,
  and Finn]{clavera2018learning}
Ignasi Clavera, Anusha Nagabandi, Simin Liu, Ronald~S. Fearing, Pieter Abbeel,
  Sergey Levine, and Chelsea Finn.
\newblock Learning to adapt in dynamic, real-world environments through
  meta-reinforcement learning.
\newblock In \emph{International Conference on Learning Representations}, 2019.
\newblock URL \url{https://openreview.net/forum?id=HyztsoC5Y7}.

\bibitem[Chen and Liu(2018)]{chen2018lifelong}
Zhiyuan Chen and Bing Liu.
\newblock Lifelong machine learning.
\newblock \emph{Synthesis Lectures on Artificial Intelligence and Machine
  Learning}, 12\penalty0 (3):\penalty0 1--207, 2018.

\bibitem[Nguyen et~al.(2018)Nguyen, Li, Bui, and Turner]{v.2018variational}
Cuong~V. Nguyen, Yingzhen Li, Thang~D. Bui, and Richard~E. Turner.
\newblock Variational continual learning.
\newblock In \emph{International Conference on Learning Representations}, 2018.
\newblock URL \url{https://openreview.net/forum?id=BkQqq0gRb}.

\bibitem[Jerfel et~al.(2019)Jerfel, Grant, Griffiths, and
  Heller]{jerfel2019reconciling}
Ghassen Jerfel, Erin Grant, Tom Griffiths, and Katherine~A Heller.
\newblock Reconciling meta-learning and continual learning with online mixtures
  of tasks.
\newblock In \emph{Advances in Neural Information Processing Systems}, pages
  9119--9130, 2019.

\bibitem[Deleu and Bengio(2018)]{deleu2018effects}
Tristan Deleu and Yoshua Bengio.
\newblock The effects of negative adaptation in model-agnostic meta-learning.
\newblock \emph{arXiv preprint arXiv:1812.02159}, 2018.

\bibitem[Nagabandi et~al.(2019)Nagabandi, Finn, and Levine]{nagabandi2018deep}
Anusha Nagabandi, Chelsea Finn, and Sergey Levine.
\newblock Deep online learning via meta-learning: Continual adaptation for
  model-based {RL}.
\newblock In \emph{International Conference on Learning Representations}, 2019.
\newblock URL \url{https://openreview.net/forum?id=HyxAfnA5tm}.

\bibitem[Vuorio et~al.(2019)Vuorio, Sun, Hu, and Lim]{vuorio2019multimodal}
Risto Vuorio, Shao-Hua Sun, Hexiang Hu, and Joseph~J Lim.
\newblock Multimodal model-agnostic meta-learning via task-aware modulation.
\newblock In \emph{Advances in Neural Information Processing Systems}, pages
  1--12, 2019.

\bibitem[Aljundi et~al.(2019)Aljundi, Kelchtermans, and
  Tuytelaars]{aljundi2019task}
Rahaf Aljundi, Klaas Kelchtermans, and Tinne Tuytelaars.
\newblock Task-free continual learning.
\newblock In \emph{Proceedings of the IEEE Conference on Computer Vision and
  Pattern Recognition}, pages 11254--11263, 2019.

\bibitem[Lee et~al.(2020)Lee, Ha, Zhang, and Kim]{Lee2020A}
Soochan Lee, Junsoo Ha, Dongsu Zhang, and Gunhee Kim.
\newblock A neural dirichlet process mixture model for task-free continual
  learning.
\newblock In \emph{International Conference on Learning Representations}, 2020.
\newblock URL \url{https://openreview.net/forum?id=SJxSOJStPr}.

\bibitem[Chua et~al.(2018)Chua, Calandra, McAllister, and Levine]{chua2018deep}
Kurtland Chua, Roberto Calandra, Rowan McAllister, and Sergey Levine.
\newblock Deep reinforcement learning in a handful of trials using
  probabilistic dynamics models.
\newblock In \emph{Advances in Neural Information Processing Systems}, pages
  4754--4765, 2018.

\bibitem[Yoon et~al.(2018)Yoon, Yang, Lee, and Hwang]{yoon2018lifelong}
Jaehong Yoon, Eunho Yang, Jeongtae Lee, and Sung~Ju Hwang.
\newblock Lifelong learning with dynamically expandable networks.
\newblock In \emph{International Conference on Learning Representations}, 2018.
\newblock URL \url{https://openreview.net/forum?id=Sk7KsfW0-}.

\bibitem[Rasmussen(2003)]{rasmussen2003gaussian}
Carl~Edward Rasmussen.
\newblock Gaussian processes in machine learning.
\newblock In \emph{Summer School on Machine Learning}, pages 63--71. Springer,
  2003.

\bibitem[Gardner et~al.(2018)Gardner, Pleiss, Weinberger, Bindel, and
  Wilson]{gardner2018gpytorch}
Jacob Gardner, Geoff Pleiss, Kilian~Q Weinberger, David Bindel, and Andrew~G
  Wilson.
\newblock Gpytorch: Blackbox matrix-matrix gaussian process inference with gpu
  acceleration.
\newblock In \emph{Advances in Neural Information Processing Systems}, pages
  7576--7586, 2018.

\bibitem[Lin(2013)]{lin2013online}
Dahua Lin.
\newblock Online learning of nonparametric mixture models via sequential
  variational approximation.
\newblock In \emph{Advances in Neural Information Processing Systems}, pages
  395--403, 2013.

\bibitem[Titsias(2009)]{titsias2009variational}
Michalis Titsias.
\newblock Variational learning of inducing variables in sparse gaussian
  processes.
\newblock In \emph{Artificial Intelligence and Statistics}, pages 567--574,
  2009.

\bibitem[Duan et~al.(2016)Duan, Schulman, Chen, Bartlett, Sutskever, and
  Abbeel]{duan2016rl}
Yan Duan, John Schulman, Xi~Chen, Peter~L Bartlett, Ilya Sutskever, and Pieter
  Abbeel.
\newblock Rl2: Fast reinforcement learning via slow reinforcement learning.
\newblock \emph{arXiv preprint arXiv:1611.02779}, 2016.

\bibitem[Wang et~al.(2016)Wang, Kurth-Nelson, Tirumala, Soyer, Leibo, Munos,
  Blundell, Kumaran, and Botvinick]{wang2016learning}
Jane~X Wang, Zeb Kurth-Nelson, Dhruva Tirumala, Hubert Soyer, Joel~Z Leibo,
  Remi Munos, Charles Blundell, Dharshan Kumaran, and Matt Botvinick.
\newblock Learning to reinforcement learn.
\newblock \emph{arXiv preprint arXiv:1611.05763}, 2016.

\bibitem[Finn et~al.(2017)Finn, Abbeel, and Levine]{finn2017model}
Chelsea Finn, Pieter Abbeel, and Sergey Levine.
\newblock Model-agnostic meta-learning for fast adaptation of deep networks.
\newblock In \emph{Proceedings of the 34th International Conference on Machine
  Learning-Volume 70}, pages 1126--1135. JMLR. org, 2017.

\bibitem[Houthooft et~al.(2018)Houthooft, Chen, Isola, Stadie, Wolski, Ho, and
  Abbeel]{houthooft2018evolved}
Rein Houthooft, Yuhua Chen, Phillip Isola, Bradly Stadie, Filip Wolski,
  OpenAI~Jonathan Ho, and Pieter Abbeel.
\newblock Evolved policy gradients.
\newblock In \emph{Advances in Neural Information Processing Systems}, pages
  5400--5409, 2018.

\bibitem[Rothfuss et~al.(2019)Rothfuss, Lee, Clavera, Asfour, and
  Abbeel]{rothfuss2018promp}
Jonas Rothfuss, Dennis Lee, Ignasi Clavera, Tamim Asfour, and Pieter Abbeel.
\newblock Pro{MP}: Proximal meta-policy search.
\newblock In \emph{International Conference on Learning Representations}, 2019.
\newblock URL \url{https://openreview.net/forum?id=SkxXCi0qFX}.

\bibitem[Peters and Schaal(2008)]{peters2008reinforcement}
Jan Peters and Stefan Schaal.
\newblock Reinforcement learning of motor skills with policy gradients.
\newblock \emph{Neural networks}, 21\penalty0 (4):\penalty0 682--697, 2008.

\bibitem[Wilson et~al.(2007)Wilson, Fern, Ray, and Tadepalli]{wilson2007multi}
Aaron Wilson, Alan Fern, Soumya Ray, and Prasad Tadepalli.
\newblock Multi-task reinforcement learning: a hierarchical bayesian approach.
\newblock In \emph{Proceedings of the 24th international conference on Machine
  learning}, pages 1015--1022, 2007.

\bibitem[Candy(1991)]{candy1991self}
Philip~C Candy.
\newblock \emph{Self-Direction for Lifelong Learning. A Comprehensive Guide to
  Theory and Practice.}
\newblock ERIC, 1991.

\bibitem[Ruder(2017)]{ruder2017overview}
Sebastian Ruder.
\newblock An overview of multi-task learning in deep neural networks.
\newblock \emph{arXiv preprint arXiv:1706.05098}, 2017.

\bibitem[Evgeniou and Pontil(2004)]{evgeniou2004regularized}
Theodoros Evgeniou and Massimiliano Pontil.
\newblock Regularized multi--task learning.
\newblock In \emph{Proceedings of the tenth ACM SIGKDD international conference
  on Knowledge discovery and data mining}, pages 109--117, 2004.

\bibitem[Jacob et~al.(2009)Jacob, Vert, and Bach]{jacob2009clustered}
Laurent Jacob, Jean-philippe Vert, and Francis~R Bach.
\newblock Clustered multi-task learning: A convex formulation.
\newblock In \emph{Advances in neural information processing systems}, pages
  745--752, 2009.

\bibitem[Jain and Neal(2004)]{jain2004split}
Sonia Jain and Radford~M Neal.
\newblock A split-merge markov chain monte carlo procedure for the dirichlet
  process mixture model.
\newblock \emph{Journal of computational and Graphical Statistics}, 13\penalty0
  (1):\penalty0 158--182, 2004.

\bibitem[Dahl(2005)]{dahl2005sequentially}
David~B Dahl.
\newblock Sequentially-allocated merge-split sampler for conjugate and
  nonconjugate dirichlet process mixture models.
\newblock \emph{Journal of Computational and Graphical Statistics}, 11\penalty0
  (6), 2005.

\bibitem[Blei et~al.(2006)Blei, Jordan, et~al.]{blei2006variational}
David~M Blei, Michael~I Jordan, et~al.
\newblock Variational inference for dirichlet process mixtures.
\newblock \emph{Bayesian analysis}, 1\penalty0 (1):\penalty0 121--143, 2006.

\bibitem[Broderick et~al.(2013)Broderick, Boyd, Wibisono, Wilson, and
  Jordan]{broderick2013streaming}
Tamara Broderick, Nicholas Boyd, Andre Wibisono, Ashia~C Wilson, and Michael~I
  Jordan.
\newblock Streaming variational bayes.
\newblock In \emph{Advances in neural information processing systems}, pages
  1727--1735, 2013.

\bibitem[Huynh et~al.(2016)Huynh, Phung, and Venkatesh]{huynh2016streaming}
Viet Huynh, Dinh Phung, and Svetha Venkatesh.
\newblock Streaming variational inference for dirichlet process mixtures.
\newblock In \emph{Asian Conference on Machine Learning}, pages 237--252, 2016.

\bibitem[Tank et~al.(2015)Tank, Foti, and Fox]{tank2015streaming}
Alex Tank, Nicholas Foti, and Emily Fox.
\newblock Streaming variational inference for bayesian nonparametric mixture
  models.
\newblock In \emph{Artificial Intelligence and Statistics}, pages 968--976,
  2015.

\bibitem[Fox et~al.(2011)Fox, Sudderth, Jordan, and Willsky]{fox2011sticky}
Emily~B Fox, Erik~B Sudderth, Michael~I Jordan, and Alan~S Willsky.
\newblock A sticky hdp-hmm with application to speaker diarization.
\newblock \emph{The Annals of Applied Statistics}, pages 1020--1056, 2011.

\bibitem[Kearns et~al.(1998)Kearns, Mansour, and Ng]{kearns1998information}
Michael Kearns, Yishay Mansour, and Andrew~Y Ng.
\newblock An information-theoretic analysis of hard and soft assignment methods
  for clustering.
\newblock In \emph{Learning in graphical models}, pages 495--520. Springer,
  1998.

\bibitem[Neal(2000)]{neal2000markov}
Radford~M Neal.
\newblock Markov chain sampling methods for dirichlet process mixture models.
\newblock \emph{Journal of computational and graphical statistics}, 9\penalty0
  (2):\penalty0 249--265, 2000.

\bibitem[Rasmussen and Ghahramani(2002)]{rasmussen2002infinite}
Carl~E Rasmussen and Zoubin Ghahramani.
\newblock Infinite mixtures of gaussian process experts.
\newblock In \emph{Advances in neural information processing systems}, pages
  881--888, 2002.

\bibitem[Tran et~al.(2015)Tran, Ranganath, and Blei]{tran2015variational}
Dustin Tran, Rajesh Ranganath, and David~M Blei.
\newblock The variational gaussian process.
\newblock \emph{arXiv preprint arXiv:1511.06499}, 2015.

\bibitem[Titsias et~al.(2019)Titsias, Schwarz, Matthews, Pascanu, and
  Teh]{titsias2019functional}
Michalis~K Titsias, Jonathan Schwarz, Alexander G de~G Matthews, Razvan
  Pascanu, and Yee~Whye Teh.
\newblock Functional regularisation for continual learning with gaussian
  processes.
\newblock In \emph{International Conference on Learning Representations}, 2019.

\bibitem[Guha et~al.(2019)Guha, Ho, and Nguyen]{guha2019posterior}
Aritra Guha, Nhat Ho, and XuanLong Nguyen.
\newblock On posterior contraction of parameters and interpretability in
  bayesian mixture modeling.
\newblock \emph{arXiv preprint arXiv:1901.05078}, 2019.

\bibitem[Kim et~al.(2019)Kim, Mnih, Schwarz, Garnelo, Eslami, Rosenbaum,
  Vinyals, and Teh]{kim2018attentive}
Hyunjik Kim, Andriy Mnih, Jonathan Schwarz, Marta Garnelo, Ali Eslami, Dan
  Rosenbaum, Oriol Vinyals, and Yee~Whye Teh.
\newblock Attentive neural processes.
\newblock In \emph{International Conference on Learning Representations}, 2019.
\newblock URL \url{https://openreview.net/forum?id=SkE6PjC9KX}.

\bibitem[Qin et~al.(2019)Qin, Zhu, Qin, Wang, and Zhao]{qin2019recurrent}
Shenghao Qin, Jiacheng Zhu, Jimmy Qin, Wenshuo Wang, and Ding Zhao.
\newblock Recurrent attentive neural process for sequential data.
\newblock \emph{arXiv preprint arXiv:1910.09323}, 2019.

\bibitem[Teh(2010)]{teh2010dirichlet}
Yee~Whye Teh.
\newblock Dirichlet process., 2010.

\bibitem[Galashov et~al.(2019)Galashov, Schwarz, Kim, Garnelo, Saxton, Kohli,
  Eslami, and Teh]{galashov2019meta}
Alexandre Galashov, Jonathan Schwarz, Hyunjik Kim, Marta Garnelo, David Saxton,
  Pushmeet Kohli, SM~Eslami, and Yee~Whye Teh.
\newblock Meta-learning surrogate models for sequential decision making.
\newblock \emph{arXiv preprint arXiv:1903.11907}, 2019.

\bibitem[Garnelo et~al.(2018)Garnelo, Rosenbaum, Maddison, Ramalho, Saxton,
  Shanahan, Teh, Rezende, and Eslami]{pmlr-v80-garnelo18a}
Marta Garnelo, Dan Rosenbaum, Christopher Maddison, Tiago Ramalho, David
  Saxton, Murray Shanahan, Yee~Whye Teh, Danilo Rezende, and S.~M.~Ali Eslami.
\newblock Conditional neural processes.
\newblock In Jennifer Dy and Andreas Krause, editors, \emph{Proceedings of the
  35th International Conference on Machine Learning}, volume~80 of
  \emph{Proceedings of Machine Learning Research}, pages 1704--1713,
  Stockholmsmässan, Stockholm Sweden, 10--15 Jul 2018. PMLR.
\newblock URL \url{http://proceedings.mlr.press/v80/garnelo18a.html}.

\bibitem[Brockman et~al.(2016)Brockman, Cheung, Pettersson, Schneider,
  Schulman, Tang, and Zaremba]{1606.01540}
Greg Brockman, Vicki Cheung, Ludwig Pettersson, Jonas Schneider, John Schulman,
  Jie Tang, and Wojciech Zaremba.
\newblock Openai gym, 2016.

\bibitem[Todorov et~al.(2012)Todorov, Erez, and Tassa]{todorov2012mujoco}
Emanuel Todorov, Tom Erez, and Yuval Tassa.
\newblock Mujoco: A physics engine for model-based control.
\newblock In \emph{Intelligent Robots and Systems (IROS), 2012 IEEE/RSJ
  International Conference on}, pages 5026--5033. IEEE, 2012.
\newblock URL \url{https://ieeexplore.ieee.org/abstract/document/6386109/}.

\bibitem[Leurent(2018)]{highway-env}
Edouard Leurent.
\newblock An environment for autonomous driving decision-making.
\newblock \url{https://github.com/eleurent/highway-env}, 2018.

\bibitem[Albrecht and Stone(2018)]{albrecht2018autonomous}
Stefano~V Albrecht and Peter Stone.
\newblock Autonomous agents modelling other agents: A comprehensive survey and
  open problems.
\newblock \emph{Artificial Intelligence}, 258:\penalty0 66--95, 2018.

\end{thebibliography}

\clearpage

% \section{Supplementary Material}
\beginsupplement

\section*{Supplementary Material}

\section{Simulation Envirionments}
\label{section:suppl:simulations}

We present details of the three non-stationary simulation environments in this section. Since we deal with real-world physical tasks, each kind of task has its own dynamics type. The RL agent encounters different types of dynamics sequentially in our setting. The switch of dynamics is assumed to finish within a single timestep. Each type of dynamics may last for several episodes, and we call each episode as one subtask since the dynamics type within an episode is invariant. Note that our method is not restricted by the episodic assumption since the task index and task boundary are unknown. A summary of the key parameters is shown in Table 1.

\subsection{Changeable Pole Length and Mass of CartPole-SwingUp}

$\mathtt{CartPole}$-$\mathtt{SwingUp}$ consists of a cart moving horizontally and a pole with one end attached at the center of the cart. 
We modify the simulation environment based on the OpenAI Gym environment~\citep{1606.01540}. 
Different types of dynamics have different pole mass $m$ and pole length $l$.
In our setting, the agent encounters four types of dynamics sequentially with the combinations $(m,l)$ as $(0.4, 0.5)$, $(0.4, 0.7)$, $(0.8, 0.5)$, $(0.8, 0.7)$. 
Denote the position of the cart as $x$ and the angle of the pole as $\theta$. 
The state of the environment is $\xv = (x, \dot x, \cos{\theta}, \sin{\theta}, \dot \theta )$ and action $\uv$ is the horizontal force applied on the cart.
Therefore, the GP input is a 6 dimensional vector $\txv = (\xv, \uv )$. The GP target is a 5 dimensional vector as the state increment $\yv = \Delta \xv= (\Delta x, \Delta \dot x, \Delta \cos{\theta}, \Delta \sin{\theta}, \Delta \dot \theta )$.

\subsection{Varying Torso Mass in HalfCheetah}
% maybe the title need to be changed
In $\mathtt{HalfCheetah}$, we aim to control a halfheetah in flat ground and make it run as far as possible. The environment is modified based on MuJoCo~\citep{todorov2012mujoco}.
We notice that the torso mass $m$ significantly affects the running performance.
Therefore, we create two different types of dynamics by changing the torso mass $m$ iteratively between $14$ and $34$ to simulate real-world delivery situations. 
We denote the nine joints of a halfcheetah as (root\_x, root\_y, root\_z, back\_thigh, back\_shin, back\_foot, front\_thigh, front\_shin, front\_foot). The state $\xv$ is 18-dimensional consisting of each joint's position and velocity. The action $\uv$ is 6-dimensional, including the actuator actions applied to the last six physical joints. Therefore, the GP input $\txv= (\xv, \uv)$ is a 24-dimensional, and the GP target is the 18-dimensional state increment $y = \Delta \xv$.

\subsection{Dynamic Surrounding Vehicles in Highway-Intersection}
% $\mathtt{Highway}$: 
The $\mathtt{Highway}$-$\mathtt{Intersection}$ environment is modified based on highway-env~\citep{highway-env}.
We adapt the group modeling~\citep{albrecht2018autonomous} concept and treat all the other surrounding vehicles' behaviors as part of the environment dynamics. In addition to modeling the interactions (analogous to the effect of ego vehicle action to other vehicles), the mixture model also needs to learn the ego vehicles' dynamics (analogous to the action's effect on ego vehicle).
For simplicity, we only experiment with environments containing only one surrounding vehicle. 
Note that the multi-vehicle environment can be generated by following the same setting but requires modification of pipelines. For example, to evaluate the posterior probability of all surrounding vehicles from GP components, we can query the mixture model multiple times and then calculate the predictive distribution for each vehicle.
%Note that the multi-vehicle setting can be generated following the same representation, but requires modification of pipelines such as querying the mixture model multiple times to calculate the predictive distribution for each surrounding vehicle and averaging the posterior probability of each surrounding vehicle given GP components. 

We consider an intersection with a two-lane road. 
The downside, left, upside, and right entry is denoted with index 0,1,2,3, respectively.
The ego vehicle $A_0$ has a fixed initial state as the right lane of entry 0 and a fixed destination as the right lane of entry 1. In other words, $A_0$ tries to do a left turn with high velocity and avoid collision with others.
Each type of dynamics has different start and goal positions of the surrounding vehicle $A_1$. $A_0$ encounters three types of interactions with the combinations of $A_1$'s start entry and goal entry as $(2,1)$, $(2,0)$ and $(1,2)$. Note that when $A_1$ emerges at entry 2 and heading to entry 0, $A_0$ faces a typical unprotected left turn scenario.

Denote the positions and heading of a vehicle as $(x,y,h)$. The state of $A_0$ is $(x, y, \dot{x}, \dot{y}, \cos{h}, \sin{h})$ in the word-fixed frame. The state of $A_1$ is $(x_{rel}, y_{rel}, \dot{x}_{rel}, \dot{y}_{rel}, \cos{h}_{rel}, \sin{h}_{rel})$ evaluated in the body frame fixed at $A_0$. We directly control the ego vehicle $A_0$'s acceleration $a$ and steering angle $\theta$. Therefore, the GP input $\txv$ is a 14 dimensional vector consisting of the $A_0$'s state and action as well as $A_1$'s state. The GP target is the increment of the ego vehicle's and the other vehicle's states.

\begin{table}[t]
    \centering
    \caption{Simulation Environment Details. Each type of dynamics last for 3 episodes.}
    \begin{tabular}{l c c c} % p{0.25\textwidth} 
        \hline
        Environment & $\mathtt{CartPole}$ & $\mathtt{HalfCheetah}$ & $\mathtt{Intersection}$ \\
        \hline
        State Dimension & 5 &  18 & 12 \\
        Action Dimension & 1 & 6 &  2 \\
        Episode Length & 200 & 200 & 40 \\
        Simulation Interval (s) & 0.04 & 0.01 & 0.1\\
        Early Stop  & True when $x$ out of limit & False & True when Collision \\
        No. episodes / Dynamics & 3 & 3 & 3 \\ \hline
    \end{tabular}
    \label{tab:envdims}
\end{table}

\begin{table}[t]
    \centering
    \caption{Model Parameters.}
    \begin{tabular}{l c c c}
        \hline
        Parameter & $\mathtt{CartPole}$ & $\mathtt{HalfCheetah}$ & $\mathtt{Intersection}$\\ 
        \hline
        concentration parameter $\alpha$ & 0.1 & 1.5 & 0.5 \\ 
        sticky paramter $\beta$ & 1 & 1 & 1 \\
        initial noise $\sigma_i, i=1,...c$ & 0.001 & 0.1 & 0.001 \\
        initial output scale $w_i, i=1,...c$ & 0.5 & 10.0 & 0.5 \\
        initial lengthscale $1/w_{i,j}, i,j=1,...c$ & 1.0 & 1.0 & 1.0 \\
        merge KL threshold $\epsilon$ & 20 & 10 & 70 \\
        merge trigger $n_{merge}$ & 15 & 5 & 10 \\
        data distillation trigger $n_{distill}$ & 1500 & 2000 & 1500 \\
        inducing point number $m$ & 1300 & 1800 & 1300 \\
        GP update Steps / timestep & 10 & 5 & 10 \\
        learning rate  & 0.1 & 0.1 & 0.1 \\
        discount $\gamma$ & 1 & 1 & 1\\
        MPC plan horizion & 20 & 15 & 20 \\ 
        CEM popsize & 200 & 200 & 200 \\
        CEM No. elites & 20 & 10 & 20 \\
        CEM iterations & 5 & 5 & 5\\ 
        \hline
    \end{tabular}
    
    \label{tab:GPMM_param}
\end{table}

\section{Method Details}
\label{section:suppl:parameters}

The computing infrastructure is a desktop with twelve 64-bit CPU (model: Intel(R) Core(TM) i7-8700K CPU @ 3.70GHz) and a GPU (model: NVIDIA GeForce RTX 2080 Ti). Since our method is pre-train free, we do not need to collect data from different dynamics beforehand. However, for most of the baselines, we collect data to pre-train them as detailed in Section~\ref{section:suppl:param_baseline}.
In our online setting, there is no clear boundary between training and testing. More concretely, at each time step, our model is updated by the streaming collected data and evaluated in MPC to select the optimal action.

\subsection{Parameters of Our Method}

We list key parameters of our proposed method in Table~\ref{tab:GPMM_param}. The concentration parameter $\alpha$ controls the generation of new GP components. The larger the $\alpha$, the more likely a new GP component is spawned. The sticky parameter $\beta$ increases the self-transition probability of each component. The initial parameter $\thetav_0$ for a new GP consists of initial noise $\sigma_i$, initial output scale $w_i$ and initial lengthscale $1/w_{i,j}$. Larger lengthscale and output scale help alleviate the overfitting problem by increasing the effect from the away points. $\mathtt{HalfCheetah}$ has a higher state dimension than the other two and thus has larger $n_{distill}$ and $m$. The planning horizon of $\mathtt{Intersection}$ is half of the total episode length since the MPC needs to do predictive collision check to achieve safe navigation. For the hyperparameter selection, we randomly search in a coarse range first and then do a grid search in a smaller hyperparameter space.

\subsection{Parameters of Baselines}
\label{section:suppl:param_baseline}
The critical parameters of using a DNN, an ANP~\citep{kim2018attentive, qin2019recurrent}, and a MAML~\citep{finn2017model} are shown in Table~\ref{tab:baseline_param}. 
The parameters for a single GP baseline and the concentration parameter $\alpha$ of the DPNN baseline are the same as the corresponding ones of our proposed method, as in Table~\ref{tab:GPMM_param}. The DNN parameters in DPNN baseline is the same as the parameters of a single DNN, as in Table~\ref{tab:baseline_param}. Note that except for the baseline using a single GP as dynamics models, all the other baselines require to collect data to pre-train the model. In our setting, we collect ten episodes from each type of dynamics as the pre-train dataset. The parameters for baselines are all carefully selected to achieve decent and equitable performance in our nonstationary setting. For instance, since $\mathtt{HalfCheetah}$ has a larger state dimension than the other two, it has larger units per layer in DNN and latent dimension in ANP.

To adapt the MAML method, in addition to the pre-train free assumption, we further release the assumption that the task boundaries between different dynamics are unknown during the pre-train procedure.
Note that MAML is pre-trained with the same amount of data as the other baselines to guarantee a fair comparison. The performance of MAML may increase if collecting more data. During the online testing period, the adapt model copies meta-model's weights and updates its weights with recently collected data at each timestep.

\begin{table}[t]
    \centering
    \caption{Parameters of the DNN, ANP and MAML Baselines. }
    \begin{tabular}{c p{0.30\textwidth} c c c}
        \hline
        Baseline & Parameter & $\mathtt{CartPole}$ & $\mathtt{HalfCheetah}$ & $\mathtt{Intersection}$\\
        \hline
         & pre-train episodes / dynamics  & 10 & 10 & 10\\ \hline
        \multirow{6}{1cm}{DNN}  & gradient steps & 200 & 100  & 100   \\
        &optimizer & Adam & Adam & Adam \\
        &learning rate & 0.0005 & 0.001 & 0.008 \\
        &hidden layers & 2 & 2 & 2 \\ 
        &units per layer & 256 & 500 & 256 \\
        &minibatch size & 512 & 256 & 128 \\ \hline
        \multirow{8}{1cm}{ANP}  & gradient steps & 200 & 800 & 100 \\
        &optimizer & Adam & Adam & Adam \\
        &learning rate & 0.0005 & 0.001 & 0.0005 \\
        & hidden layers & [256, 128, 64] & [512, 256, 128] & [256, 128, 64] \\
        & minibatch size & 1024 & 64 & 1024  \\ 
        & context number & 100 & 100 & 100 \\
        & target number & 25 & 25 & 25\\
        & latent dimension & 64 & 128 & 64 \\ \hline
        \multirow{8}{1cm}{MAML} & hidden layers & 2 & 2 & 2 \\
        & units per layer & 500 & 500 & 500 \\
        & step size $\alpha$ & 0.01 & 0.01 & 0.01 \\
        & step size $\beta$ & 0.001 & 0.001 & 0.001 \\
        & meta steps & 200 & 100 & 100 \\
        & adapt learning rate & 0.001 & 0.001 & 0.001 \\
        & adapt step & 10 & 10 & 10  \\
        & meta batch size & 1 & 1 & 1 \\ \hline
    \end{tabular}
    \label{tab:baseline_param}
\end{table}

\section{Additional Experiment Results}
\label{section:suppl:expresults}

\subsection{Dynamics Assignments with Dirichlet Process Prior}
\label{section:suppl:assignments}

\begin{figure}[t]
\begin{subfigure}{.5\textwidth}
  \centering
  % include first image
  \includegraphics[width=1.\linewidth]{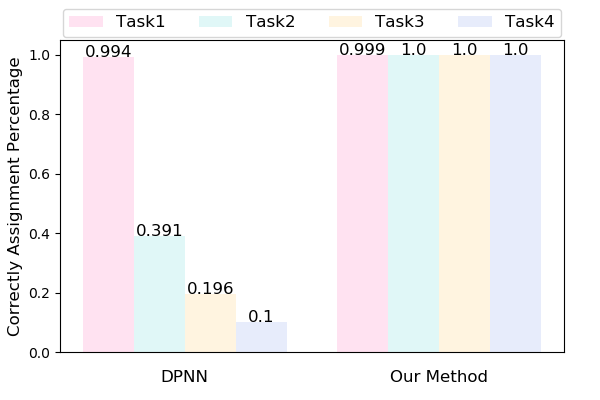}  
  \caption{$\mathtt{CartPole}$-$\mathtt{Sqingup}$}
  \label{fig:sub-first}
\end{subfigure}
\begin{subfigure}{.5\textwidth}
  \centering
  % include second image
  \includegraphics[width=1\linewidth]{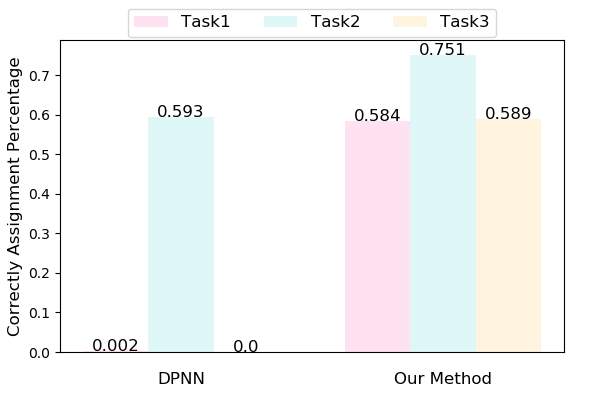}  
  \caption{$\mathtt{Highway}$-$\mathtt{Intersection}$}
  \label{fig:sub-second}
\end{subfigure}
\caption{Correct Assignment Percentage using DPNN and our method. 
% Each bar is evaluated with 10 runs. 
Our method ourperforms DPNN in terms of task assignments.}
\label{fig:assignment_results_supple}
\end{figure}

We show that using pure DP prior is not sufficient to capture the dynamics assignments of streaming data in real-world physical tasks by visualizing the cluster results in $\mathtt{CartPole}$-$\mathtt{SwingUp}$, as in Figure~\ref{fig:cluster_cp_DP}. In this section, we show more statistics about the dynamics assignments with DP prior by comparing the performance of DPNN and our method in Figure~\ref{fig:assignment_results_supple}. 

In $\mathtt{CartPole}$-$\mathtt{Sqingup}$, we can see that our method can accurately detect the dynamics shift and cluster the streaming data to the right type of dynamics. However, when using DPNN, the more types of dynamics encountered, the less accurate the assignments are. 
In $\mathtt{Highway}$-$\mathtt{Intersection}$, our method sometimes cluster the data points into the wrong dynamics. We hypothesize that this may be due to the overlap in the spatial domain of different interactions. However, our method still outperforms the DPNN in terms of dynamics assignments. DPNN can only stably identify the second task, and either frequently generate new clusters for the first and third task or identify them as the second task. We notice that the clustering accuracy of DPNN heavily relies on the number of previous states concatenated (the length of the short term memory)~\citep{nagabandi2018deep}. To make the dynamics assignment of DPNN more stable, in our setting, we use 50 (1/4 episode length) previous data points to determine the dynamics assignment in $\mathtt{CartPole}$ and 20 (1/2 episode length) in $\mathtt{Intersection}$.

\begin{table}[t]
    \centering
    \caption{Reward Mean and Standard Deviation (std). The bold numbers indicate the maximum means and minimum stds in each task. }
    \begin{tabular}{ p{0.06\textwidth} c c c c c c c c c c}
        \hline
        &  & \multicolumn{4}{c}{$\mathtt{CartPole}$} & \multicolumn{2}{c}{$\mathtt{HalfCheetah}$} & \multicolumn{3}{c}{ $\mathtt{Intersection}$}\\
        &  & Task1 & Task2 & Task3 & Task4 & Task1 & Task2 & Task1 & Task2 & Task3\\ \hline
        
         \multirow{2}{0cm}{Our Method} & mean & 177.62 & \textbf{183.10} & \textbf{181.37} & \textbf{182.54} & \textbf{36.11} & \textbf{34.14} & 60.66 & \textbf{54.77} & \textbf{52.48} \\
         & std & 6.61 & \textbf{3.29} & \textbf{1.86} & \textbf{1.10} & \textbf{9.22} & 13.05 & 1.29 & \textbf{1.80} & \textbf{0.77}  \\ \hline
         
         \multirow{2}{0.5cm}{GP}&mean & 178.04& 169.46& 174.45& 171.67& 2.70& -21.26& 53.73& 39.23& 44.25 \\
         & std & \textbf{3.40} & 7.39& 8.94& 6.12& 14.70& 24.49& 6.56& 20.31& 6.91 \\ \hline
         
         \multirow{2}{0.5cm}{DNN}&mean & 176.57& 164.72& 177.74& 164.21& 28.89& 0.53& 43.42& 2.31& 46.62 \\
         & std & 8.14& 16.41& 5.21& 7.24& 35.02& 15.60& 9.73& 7.34& 6.12 \\ \hline
         
         \multirow{2}{0.5cm}{ANP}&mean & 175.33& 170.74& 171.29& 160.87& -4.27& -20.70& \textbf{62.68}& 44.04& 36.91  \\
         & std & 6.61& 4.44& 6.68& 8.34& 16.80& 22.09& \textbf{0.75}& 17.95& 19.86 \\ \hline
         
         \multirow{2}{0.5cm}{MAML}&mean & 144.91& 150.80& 127.55& 134.36& -70.49& -51.32&  39.29& 39.14& 40.31 \\
         & std & 28.22& 18.74& 32.82& 13.95& 36.57& 12.13& 3.36& 4.21& 5.89 \\ \hline
         
         \multirow{2}{0.5cm}{DPNN}&mean & \textbf{187.24} & 182.52& 180.20& 161.58& -42.88 & -39.30 & 59.97& 46.89& 18.64 \\
         & std & 5.03& 6.82& 3.09& 45.27& 16.37& \textbf{8.14} & 2.92& 15.33& 22.42\\ \hline
    \end{tabular}
    \label{tab:rewards}
\end{table}

\subsection{Will Dynamics Assignments Affect Task Performance?}
\label{section:suppl:performances}

To investigate whether the correct dynamics assignments improve the task performance, we compare the accumulated subtask rewards of our method, the single GP baseline, and the DPNN baseline. 
The single GP baseline is the ablation version of our method without dynamics assignments. 
In other words, using a single GP indicates clustering different dynamics into a single cluster.
The DPNN has less accurate dynamics assignments than our method, as detailed in Section~\ref{section:suppl:assignments}.

Table~\ref{tab:rewards} and Figure~\ref{fig:cartpole-subtask-rewards} show that our method has higher rewards and smaller variances than the baselines in most situations. Since our method performs better than the single GP baseline in all three nonstationary environments, it shows that the dynamics recognition help increase the task performance. 
Note that DPNN has higher rewards than our method in the first task of $\mathtt{CartPole}$-$\mathtt{SwingUp}$. We hypothesize that this may be due to the pre-train procedure of DPNN. However, our method outperforms DPNN in all the other dynamics, which indicates that accurate dynamics assignments help improve task performances.

\section{Model Derivation}

In this part, we highlight the derivation of the soft assignment $\rho_n(z_{nk})$ in \ref{eq:softassignment} that was originally derived in~\citep{tank2015streaming} Eq.12-16 to ease reading. Note that we highly recommend reading~\citep{tank2015streaming} for thorough understanding. The posterior distribution $p_n(z_{0:n}, \thetav | \mathcal{D})$ for the assignments $z_{0:n}$ and model parameter $\thetav$ can be written as product of factors
\begin{align}
    p_n(z_{0:n}, \thetav | \mathcal{D}) 
    &\propto p((\txv_n, \yv_n) | \thetav_{z_n}) p(z_n | z_{0:n-1}) p_{n-1}(z_{0:n-1}, \thetav | \mathcal{D}_{0:n-1}) \nonumber \\
    &\propto p(\thetav) \prod_{i=i}^{n}  p( (\txv_i, \yv_i) | \thetav_{z_i} ) p(z_i | z_{1:i-1}) \label{seq:streamingBayes}
\end{align}
Since \ref{seq:streamingBayes} is in factorized form, we can apply Assumed Density Filtering (ADF) to approximate the posterior of the first 0 to n data pair with $\hat{q}_n(z_{0:n}, \thetav) = \prod_{k=0}^{\infty} \gamma_n(\thetav_k) \prod_{i=0}^{n} \rho_n(z_i)$. The approximate posterior for the $n+1$-th data pair is thus
\begin{align}
    &\hat{p}_{n+1}(z_{0:n+1},\thetav| \mathcal{D}_{0:n+1}) \propto p((\txv_{n+1}, \yv_{n+1})  | \thetav) p(z_{n+1}| z_{0:n})\hat{q}_n(z_{0:n}, \thetav) \\
    &\hat{q}_{n+1}(z_{0:n+1}, \thetav) = \argmin_{q_{n+1} \in \mathcal{Q}_{n+1}}KL\big(\hat{p}_{n+1}(z_{0:n+1},\thetav| \mathcal{D}_{0:n+1})  \| q_{n+1}(z_{0:n+1}, \thetav) \big)
\end{align}

With the mean field assumption, the optimal distribution for the new observation is given in \ref{eq:softassignment}.

\end{document}